%% file: main.tex
\documentclass{article}
\usepackage{ijcai22}
\usepackage{xcolor}
\usepackage{times}
\usepackage{soul}
\usepackage{url}
\usepackage[hidelinks]{hyperref}
\usepackage[utf8]{inputenc}
\usepackage[small]{caption}
\usepackage{graphicx}
\usepackage{amsmath}
\usepackage{amsthm}
\usepackage{booktabs}
\usepackage{algorithm}
\usepackage{algorithmic}
\usepackage{bbm}
\usepackage{amssymb}
\usepackage{epstopdf}
\usepackage{multirow}
\usepackage{subcaption}

\title{Explainability in Graph Neural Networks: An Experimental Survey}

\author{
Peibo Li\and
Yixing Yang\and
Maurice Pagnucco\And
Yang Song\footnote{Contact Author}\\
\affiliations
School of Computer Science and Engineering, University of New South Wales, Sydney, Australia
\\
\emails
\{peibo.li, yixing.yang, morri, yang.song1\}@unsw.edu.au
}

\begin{document}
\maketitle
\begin{abstract}
Graph neural networks (GNNs) have been extensively developed for graph representation learning in various application domains. However, similar to all other neural networks models, GNNs suffer from the black-box problem as people cannot understand the mechanism underlying them. To solve this problem, several GNN explainability methods have been proposed to explain the decisions made by GNNs. In this survey, we give an overview of the state-of-the-art GNN explainability methods and how they are evaluated. Furthermore, we propose a new evaluation metric and conduct thorough experiments to compare GNN explainability methods on real world datasets. We also suggest future directions for GNN explainability.  
\end{abstract}
\input{introduction}
\input{explainability}

\input{reviews}

\input{experiments}

\input{future}

\input{conclusion}
\clearpage
%% The file named.bst is a bibliography style file for BibTeX 0.99c
\bibliographystyle{named}
\bibliography{ijcai22}

\end{document}

%% file: introduction.tex
\section{Introduction}

Even though deep learning models have achieved unprecedented success, 
they are criticized as being \textit{black-boxes}
\cite{arrieta2020explainable}, due to the lack of explainability, where people cannot understand how these models make decisions.  
To address this issue, many different methods and frameworks have been proposed to explain Convolutional Neural Networks (CNNs). 
To name a few, Grad-CAM \cite{selvaraju2017grad} provides visual explanations by generating heat maps of the gradients from the last covolutional layer. 
LIME \cite{ribeiro2016should} generates locally faithful explanations of any classifier or regressor by learning an interpretable model locally. \cite{DBLP:conf/nips/ChenLTBRS19} proposes a prototypical part network (ProtoPNet) that is able to identify prototypical parts in images without sacrificing model performance.

Graph Neural Networks (GNNs) have been shown to be very effective for many applications, such as computer vision \cite{bai2021discriminative,xiang2021walk,zhang2021exploiting}, citation networks analysis ~\cite{gong2019exploiting,DBLP:conf/iclr/VelickovicCCRLB18}, recommendation in social networks ~\cite{fan2019graph,chaudhary2019anomaly,guo2020deep}, drug discovery \cite{rathi2019practical,xiong2019pushing}, 
and community discovery \cite{liu2019community,fang2020effective}. 
To explain GNNs, classic explainability methods for CNNs can be extended for GNNs. 
For example, similar to Grad-CAM, in \cite{pope2019explainability}, a gradient-based explainability method is proposed for explaining GNNs. 
More recently, some explainability methods focusing on the nature of the GNNs' structures have emerged.
GNNExplainer \cite{ying2019gnnexplainer} was proposed to explain GNNs by masking unimportant edges/node features. 
Specifically, GNNExplainer learns a trainable mask of the input graph for each individual sample. 
Following the masking idea, PGExplainer \cite{luo2020parameterized}, PGMExplainer \cite{vu2020pgm} and GraphMask \cite{schlichtkrull2021interpreting} were proposed to learn stand-alone models to predict important edges, with PGExplainer and GraphMask based on multilayer perceptrons and PGMExplainer based on interpretable probabilistic graphical models. 
In addition, SubgraphX \cite{pmlr-v139-yuan21c} adopts Monte Carlo Tree Search \cite{silver2017mastering} to identify important subgraphs. 
XGNN \cite{DBLP:conf/kdd/YuanTHJ20} provides a model-level explanation by training a GNN model with a graph generator which is guided by reinforcement learning. 

In this paper, we review the recent development of GNN explainability methods and provide critical analysis. 
Furthermore, to compare GNN explainability methods on any given dataset, we propose a novel human-free evaluation metric. 
To the best of our knowledge, we are the first to perform objective and thorough experiments on GNN explainability methods. 
Major limitations of existing studies and our contributions can be summarized as follows: 
1) Reviews of GNN explainability are very limited. 
The existing survey papers for GNNs \cite{wu2020comprehensive,DBLP:journals/aiopen/ZhouCHZYLWLS20,DBLP:journals/tkde/ZhangCZ22,zhang2019graph} only include a brief discussion of GNN explainability methods. The only review paper for GNN explainability \cite{yuan2020explainability} lacks an experimental study. 
Motivated by these, we present in this paper a critical review of the current state-of-the-art in GNN explainability with in-depth experimental evaluation.
2) Existing evaluation metrics require human knowledge for ground truth and thresholding, which can only be used in certain domains. To enable comparison across most of the real-world datasets, we propose a human-free evaluation metric. 
3) Existing experimental studies have only been performed on synthetic data or small molecule datasets. 
%To enhance trustworthiness, where explainability is an important aspect, in real-world tasks, 
However, GNN explainability methods should be tested on more real-world datasets in different domains, since explainability is essential to ensure the trustworthiness and transparency of GNN models, which is a critical requirement for real-world applications. Therefore, based on existing and our novel evaluation metrics, we benchmark the existing explainability methods on three citation networks. 
4) Existing evaluations for explainability methods have only been performed over the basic GCN model \cite{kipf2016semi}. 
The effectiveness of these methods over more advanced GNN models needs to be verified. 
In our work, we experimentally study explainability for different methods over multiple advanced GNN models with different aggregation functions and deeper structures.

The rest of this article is organized as follows. Section 2 provides an introduction to explainability in general. Section 3 discusses the literature of GNN explainability methods. Section 4 introduces the experiments that we conducted in detail. Section 5 discusses the future direction of GNN explainability. Section 6 summarizes this article.

%% file: explainability.tex
\section{Explainability}

Generally, explainability is one of the two most important branches of Explainable AI (XAI); the other one being interpretability. 
\textit{Interpretability} here refers to designing models with human understandable structures \cite{rudin2019stop,freitas2014comprehensible,huysmans2011empirical}, such like decision trees. 
For \textit{explainability}, it is the ability to explain a black-box model, i.e., to make the black-box model easier to understand or reveal the reason for its effectiveness, such as highlighting an area of an image that is accountable for the prediction. 
We can regard interpretability and explainability as two opposite translations, where interpretability is to translate human-understandable knowledge to machine learning models, and explainability is to translate black-box models to human-understandable knowledge. 
We refer the interested reader to \cite{guidotti2018survey} and \cite{arrieta2020explainable} for a more detailed introduction to XAI.

In this paper, we focus on explainability. Generally in deep learning, explainability methods can be grouped into two categories. The first category is \textbf{feature visualization}. 
For instance, in the image domain, saliency maps \cite{selvaraju2017grad,shrikumar2017learning,DBLP:conf/nips/LundbergL17,woo2018cbam} are used to highlight important image areas for predictions. 
In the text domain, heatmaps \cite{karpathy2015visualizing,kadar2017representation,DBLP:journals/corr/BahdanauCB14,mullenbach2018explainable} are used to visualize input-output alignment and to highlight important words in input-text. 
Despite feature visualization being very popular in image and text domains, it is not applicable for explaining graphs. 
This is because graphs provide non-Euclidean data, which are inherently hard to visualize. 
Besides, there are some drawbacks of visualizations; first, the quality of visualization is very subjective and difficult to be objectively evaluated; 
second, these methods can only be applied to limited examples, without obtaining a global view of the entire dataset. 

Another category of explanation methods is to \textbf{mimic the original models' behaviors}.  
One popular method is to train a surrogate model, which is based on inherently interpretable models (e.g., decision trees, support vector machines) and trained to approximate the black-box model's behaviors \cite{ribeiro2016should}. 
Another way is based on perturbation \cite{luo2020parameterized,schlichtkrull2021interpreting,vu2020pgm,ying2019gnnexplainer,pmlr-v139-yuan21c,DBLP:conf/kdd/YuanTHJ20}. 
Specifically, different perturbations of input features or model structures are generated, and the corresponding model behaviors are monitored and analyzed. 
In this process, those input features and model structures that decisively affect the model behaviors can be detected. 
While explanation models are less intuitive to humans, they are easier to be quantified and statistically analyzed.

To evaluate the explanation methods, there are mainly two types of evaluation criterion: \textit{Plausibility} and \textit{Faithfulness}.
\textbf{Plausibility} refers to how convincing the explanations are to human, therefore it is more related to human evaluation, which most of the existing studies rely on \cite{selvaraju2017grad,shrikumar2017learning,DBLP:conf/nips/LundbergL17,woo2018cbam}. 
However, human evaluation is not always practical and can be very subjective. 
For more objective evaluation, \textbf{Faithfullness} was proposed and widely used \cite{lakkaraju2019faithful,wiegreffe2019attention,DBLP:conf/acl/JacoviG20,schlichtkrull2021interpreting}. Faithfulness refers to whether an explanation faithfully reflects the underlying reasoning process of a model, and several evaluation metrics for faithfulness have been proposed,
including disagreement between labels assigned by the explanation model and the black-box model \cite{lakkaraju2019faithful}; 
Total Variation Distance between the adversarial trained attention model's predictions and the original attention model's predictions \cite{wiegreffe2019attention};
and, the difference of downstream task accuracy between the explanation model and the original model \cite{schlichtkrull2021interpreting}.
% All these evaluation metrics can also be redefined as \textit{fidelity}, which evaluates the degree that an explanation mimics the original model's decision based on the percentage of matches \cite{guidotti2018survey,lakkaraju2019faithful}.

%% file: reviews.tex
\section{GNN Explainability}
% GNNs \cite{GRLBook} are neural networks that capture the structure of graphs via message passing to learn node representations for downstream tasks. 
% Mathematically, a GNN is usually constructed by two parts: message passing and aggregation, which can be formulated as follows:
% \begin{align}
% &m_{u,v}^l = MESSAGE^{(l)}\big(h_u^{(l - 1)}, h_v^{(l - 1)}\big)
% &\\h_{v}^{(l)}&=AGGREGATE^{(l)}\big(\{m_{u,v}^l: u \in \mathcal{N}(v)\}\big)
% \end{align}
% where $m_{u,v}^l$ is the message passed from node $u$ to $v$, which are connected by an edge, and $h_v^{(l)}$ is the node representation for $v$ in the $l^{th}$ layer. Here, $MESSAGE$ and $AGGREGATE$ can be arbitrary differentiable functions. 

To explain how GNNs work, many different methods have been developed. 
We present a taxonomy of these methods by categorizing them based on their origin;
non-GNN-originated methods, i.e., methods extended from other deep learning domains, and GNN-originated, i.e., methods specifically designed for GNNs.
Different from the previous GNN explainability review \cite{yuan2020explainability}, we focus primarily on GNN originated methods since they are more recent. We also review the evaluation metrics and datasets in this section. 

Usually, the explanations of GNNs are represented as important subgraphs of their computational graphs. While some of the explainability methods just output the subgraphs \cite{vu2020pgm,pmlr-v139-yuan21c,DBLP:conf/kdd/YuanTHJ20}, most of them assign an importance score to every edge of the graph; and, to get the important subgraph, thresholding based on these importance scores is needed \cite{ying2019gnnexplainer,luo2020parameterized,schlichtkrull2021interpreting}.

\subsection{Non-GNN-originated Methods}
%\subsubsection{Extension Works}
\textbf{\cite{baldassarre2019explainability}} exploited the capacibility of gradient-based and decomposition-based methods for explainability. 
For gradient-based methods, they implemented Sensitivity Analysis (SA) \cite{gevrey2003review} and Guided Backpropogation (GBP) \cite{springenberg2014striving} (previosly designed for image domain) for GNNs. For decomposition-based methods, they implemented Layer-wise Relevance Propagation (LRP) \cite{montavon2017explaining}. 
This study is the first work to focus on explainability techniques for GNNs and set the ground for later development.

 \textbf{\cite{pope2019explainability}} developed analogues for GNNs of three prominent explainability methods for classic CNNs, which are contrastive gradient-based (CG) saliency maps \cite{simonyan2014deep}, Class Activation Mapping (CAM) \cite{zhou2016learning} and Excitation Backpropagation (EB) \cite{zhang2018top}, and their variants, gradient-weighted CAM (Grad-CAM) \cite{selvaraju2017grad} and contrastive EB (c-EB) \cite{zhang2018top}. They quantitatively evaluate the performance of these methods with respect to fidelity, contrastivity and sparsity. According to their experiments, they found Grad-CAM is the most suitable among the studied methods for explanations on graphs of moderate size.

\subsection{GNN-originated Methods}
% \begin{table}
% \centering
% \begin{tabular}{lrr}
% \toprule
% Method  & Explanation Level & Task \\
% \midrule
% GNNExplainer       & Single-Instance  & NC/GC      \\

% PGExplainer    & Model  & NC/GC      \\
% GraphMASK    & Model  & NC/GC      \\
% PGMExplainer    & Single-Instance  & NC/GC      \\
% SubgraphX       & Single-Instance  & NC/GC      \\
% XGNN    & Model  & GC      \\
% \bottomrule
% \end{tabular}
% \caption{Summary of GNN-originated methods. NC refers to node classification and GC refers to graph classification.}
% \label{tab:booktabs}
% \end{table}
\textbf{GNNExplainer \cite{ying2019gnnexplainer}} is the first general, model-agnostic approach for GNN-based models on any graph-based tasks. 
Given a graph $G_o$ with node features $X_o$ and a trained GNN $\Phi$, GNNExplainer can provide explanations for any node $v$. 
It aims to identify a subgraph $G_S\subset G_o$ and the associated subset of features $X_S^F$ that are important for $\Phi$'s prediction.
Here $F$ is a binary feature selector which is a feature mask, and $X_{S}^F = X_S\odot F$.  
The objective is to maximize the mutual information between the explanations and the original model.
Mathematically, it follows the optimization framework:
\begin{equation}
\max_{G_S,F} MI(Y,(G_S,F))=H(Y)-H(Y|G=G_S,X=X_S^F)
\end{equation}
where the mutual information $MI$ quantifies the change in the probability between $Y=\Phi(G_o, X_o)$ and $Y=\Phi(G_S,X_S)$.
GNNExplainer explains GNNs based on the input graph perturbation and exploits the structural information from GNNs. 
However, it suffers from the scalability problem as the size of parameters is proportional to the size of input graphs. 
Besides, GNNExplainer only provides instance-level explanations which lack a global understanding of predictions. 
Even though it is indicated that GNNExplainer can be trained in a multi-instance manner, it is only at a theoretical level without experimental justification.

 \textbf{PGExplainer \cite{luo2020parameterized}} was later proposed to provide a global understanding of predictions made by GNNs. 
Specifically, it follows the optimization framework of GNNExplainer but relaxes the edge weights from binary variables to continuous variables in the range $(0,1)$ and adopts the reparameterization trick to efficiently optimize the objective function with gradient-based methods. 
Also, PGExplainer collectively explains predictions made by a trained model on multiple instances, which enable it to have a global view of a GNN model. 
It also improves computational efficiency by using a stand-alone model to predict the importance of all edges in the graph, which makes the parameter size independent from the graph size.

\textbf{GraphMask \cite{schlichtkrull2021interpreting}} is a post-hoc method to identify unnecessary edges. 
Different from GNNExplainer and PGExplainer, it learns a simple classifier for every edge in every layer to predict if that edge can be discarded; and the classifier is trained using the entire dataset. 
The objective function can be summarized as follows:
\begin{align}
    \max_{\lambda}\min_{\pi,b}&\sum\limits_{G, X\in Dataset}\Big(\sum\limits_{k=1}^{L}\sum\limits_{(u,v)\in\mathcal{E}}\mathbbm{1}_{[\mathbb{R}\neq 0]}(g_{\pi,b}^k(u, v))\Big)\nonumber\\
    &+\lambda\Big(D_{*}[\Phi(G,X)\vert\vert \Phi(G_{S}, X)]-\beta\Big)
\end{align}
where $g_{\pi,b}^k$ computes the importance of any edge $(u,v)$ in the $k^{th}$ layer, 
$\Phi(G, X)$ and $\Phi(G_S, X)$ denote the two outputs from the original trained GNN, and the explanation model over $G_S$, 
$\mathbbm{1}$ is the indicator function and $D_*$ is a divergence function to measure how the two outputs differ. 
Sparse relaxation \cite{louizos2017learning} is used to enable gradient-descent since the original objective is not differentiable. 
Similar to PGExpaliner, GraphMask can provide a global understanding of the trained GNN model. 
Furthermore, GraphMask provides relevant paths by giving each layer a different mask.

\textbf{PGM-Explainer \cite{vu2020pgm}} adopts the Probabilistic Graphical Model (PGM) to give a model-agnostic explainer for GNNs. 
To explain a trained GNN model $\Phi$, PGM-Explainer generates a PGM, which approximates the original prediction of $\Phi$. 
Generally, PGM-Explainer contains three major components: it first generates a set of input-output pairs, called sampled data, by perturbing the original graph; 
then, the variables selection step eliminates unimportant variables from the sampled data;
and finally, it fits the explanation Bayesian Networks for the filtered data from previous steps to generate the explanation model. 
Different from other explainability models, PGM-Explainer can illustrate the dependency among explained features. 
However, PGM-Explainer is limited to instance level explanation and the learning process of Bayesian Networks is very computationally expensive.

\textbf{SubgraphX \cite{pmlr-v139-yuan21c}} explains GNNs on a subgraph level by efficiently exploring different subgraphs with Monte Carlo Tree Search (MCTS). 
To be more specific, SubgraphX builds the MCTS by setting the input graph as the root node and each of the other nodes corresponds to a connected subgraph. 
Each edge in the MCTS denotes the graph associated with a child node and is obtained by pruning nodes from the graph associated with the parent node. The MCTS algorithm explores paths from the root to leaves guided by visiting counts and rewards. 
To evaluate the importance of subgraphs, SubgraphX forms a cooperative game with generated subgraphs as players and uses the Shapley value \cite{kuhn1953contributions} as the scoring function. 
The resultant connected subgraphs are provided as explanations, which reveal the information contained in the interactions among different nodes and edges. 
The drawback of SubgraphX is similar to PGM-Explainer: it can only provide instance-level explanations and cannot be applied to large graphs as the size of search trees increase exponentially.

\textbf{XGNN \cite{DBLP:conf/kdd/YuanTHJ20}} is a model-level explanation method for graph classification problems, providing high-level insights and a generic understanding of how GNNs work. 
In particular, for each chosen class $c_i$, it generates a graph pattern that maximizes the predicted probability of the GNN model for this class. 
This can reveal the relationships between graph patterns and the predictions of GNNs. 
Graph generation is formulated as a reinforcement learning task trained via a policy gradient method based on information from the trained GNNs, where for each step, the graph generator predicts where to add an edge into the current graph. 
However, XGNN can not be applied to node classification tasks. 
Moreover, it falls into a paradox that it uses one black-box to explain another black-box.

\subsection{Evaluation Metrics}
To evaluate GNN explainability, as we introduced before, \textit{faithfulness} is commonly assessed, which refers to how accurately the explanation reflects the true reasoning process of the model. 
Another popular criterion is \textit{plausibility} \cite{lakkaraju2019faithful}, which refers to how convincing the explanation is to humans. 
Although there is not a standard evaluation metric in this field, all the currently used metrics usually reflect one of these two aspects.

\subsubsection{Plausibility Evaluation}
\textbf{Accuracy}: Accuracy is used when the datasets contain human defined ground truth explanation patterns \cite{ying2019gnnexplainer,luo2020parameterized,DBLP:conf/kdd/YuanTHJ20,vu2020pgm}. 
Accuracy measures how well the generated explanation fits the ground truth. 
Formally, it is defined as $\frac{|\mathbf{GT}\cap\mathbf{E}|}{|\mathbf{E}|}$, where $\mathbf{GT}$ is the set of edges in the ground truth explanation and  $\mathbf{E}$ is the set of edges in the generated explanation. 
Note that accuracy is related to the size of the generated explanation; hence, it often requires users to threshold the dense explanation based on the size of ground truth explanations. 
As this needs human-defined ground truth for datasets, accuracy is not applicable to most real-world datasets.

\subsubsection{Faithfulness Evaluation}
%\textbf{Sparsity and Fidelity}: 
\textbf{Sparsity and Fidelity}: 
Sparsity is usually combined with fidelity, where sparsity measures how many redundant edges are removed from the original graph. 
A higher sparsity is preferred, as intuitively the explanation subgraph should be the smallest subgraph with all information needed to make the decision. 
Fidelity measures how faithful the explanation model is to the original GNN model. 
Formally, sparsity is defined as:
\begin{align}
Sparsity = 1-\frac{m}{M}
\end{align}
where $m$ is the size of the important subgraph (i.e., the number of edges) and $M$ is the size of the original graph. 
Fidelity is defined as:
\begin{align}
    Fidelity =& P(Y = c|G) - P(Y = c|G\setminus G_S),
    \nonumber\\& c = \mathop{\arg \max}_{c \in C} P(Y = c|G)
\end{align}
where $P$ is the probabilistic distribution output by the model, $Y$ is the prediction, $G$ is the original graph, $G_S$ is the explanation and $C$ is the set of all the classes. 
This equation evaluates the faithfulness of the explanation to the model by measuring the difference between predictions from the graphs removing the important subgraphs and the original input graphs. 
%This equation measures whether the explanations are faithfully important to the model's predictions by removing the important subgraphs from the input graphs and computing the difference between predictions. 
In practice, fidelity is computed and averaged over explanations with different sparsities.

\noindent\textbf{Inverse Fidelity}:
This evaluation metric compares the accuracy of the model using the important subgraph and the accuracy of the model using the original graph w.r.t the task over the test set \cite{schlichtkrull2021interpreting}. It is defined as:
\begin{equation}
    InvFidelity = \frac{1}{N}\sum_{N}^{i=1} (\mathbbm{1}(\hat{y_i}' = y_i) - \mathbbm{1}(\hat{y_i} = y_i))
\end{equation}
where $y_i$ is the label, $N$ is the number of samples, $\hat{y_i}$ and $\hat{y_i}'$ are the predictions from original model and explanations. 

Faithfulness evaluation metrics are motivated by understanding models' underlying reasoning process, differentiated from the reasoning process produced by explainability methods, and do not require ground truth explanations; 
hence, 
they can be applied to all tasks. 
The studies that use accuracy usually select the same number of edges with the highest importance score as the number of edges in the ground truth. As discussed above, first, ground truth is not available in most of the real-world datasets; second, the human-defined ground truth is not guaranteed to be correct as we would not need to explain the models if we already know how they reason, which makes the metric less reliable. Fidelity and inverse fidelity are usually computed over different sparsities. The problem with this is that a certain sparsity can have different impact on different data points. For instance, a 50\% sparsity might not affect the model's prediction for a node with a large neighborhood, while it might change the model's prediction significantly for a node with a small neighborhood. Therefore, it is not fair to directly compare fidelity across different data points based on the same sparsity. Further discussion is presented in Section \ref{sec:exp}.

\begin{figure*}
    \centering
    \begin{minipage}{0.97\linewidth}
        \begin{minipage}{0.32\linewidth}
            \centering
            \includegraphics[width=0.99\linewidth]{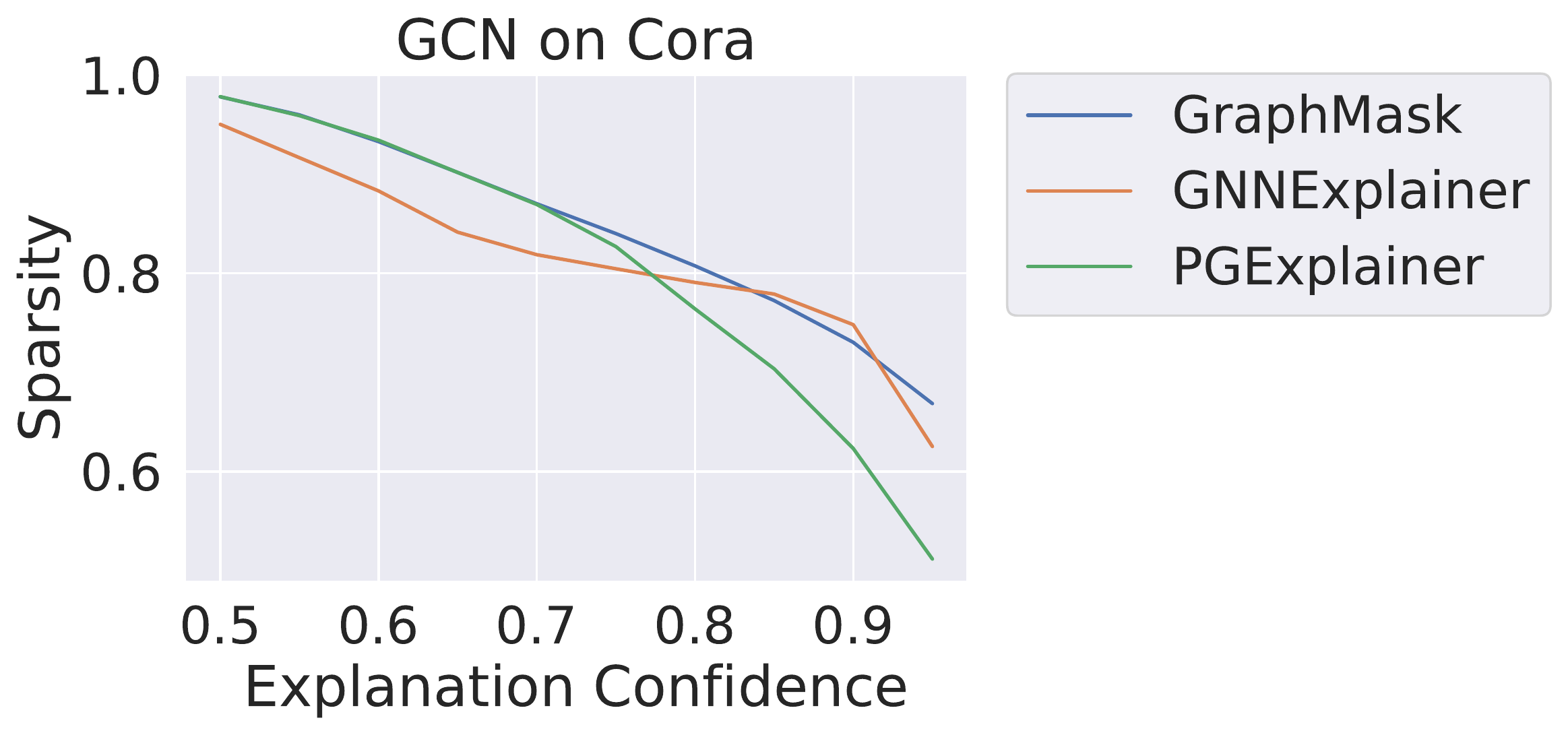}
        \end{minipage}
        \begin{minipage}{0.32\linewidth}
            \centering
            \includegraphics[width=0.99\linewidth]{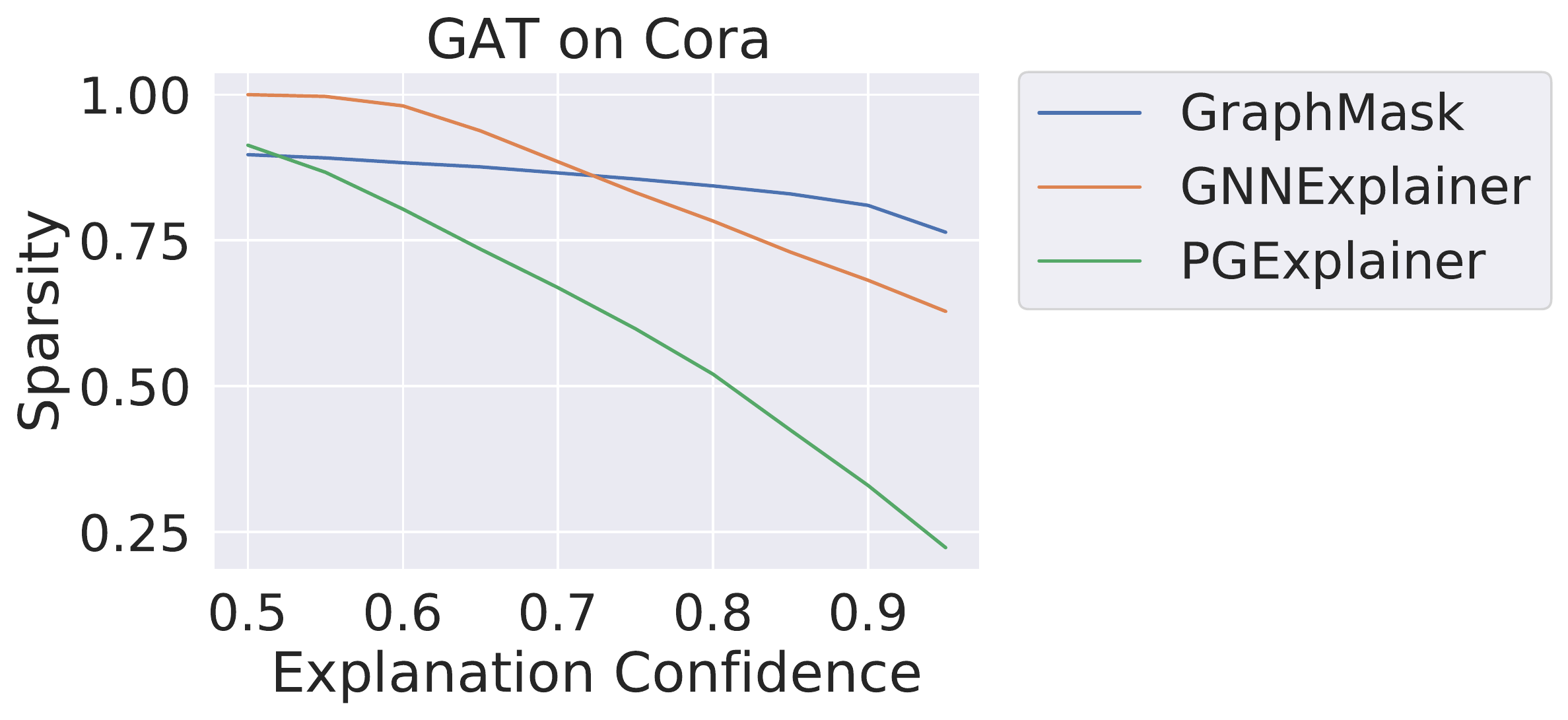}
        \end{minipage}
        \begin{minipage}{0.32\linewidth}
            \centering
            \includegraphics[width=0.99\linewidth]{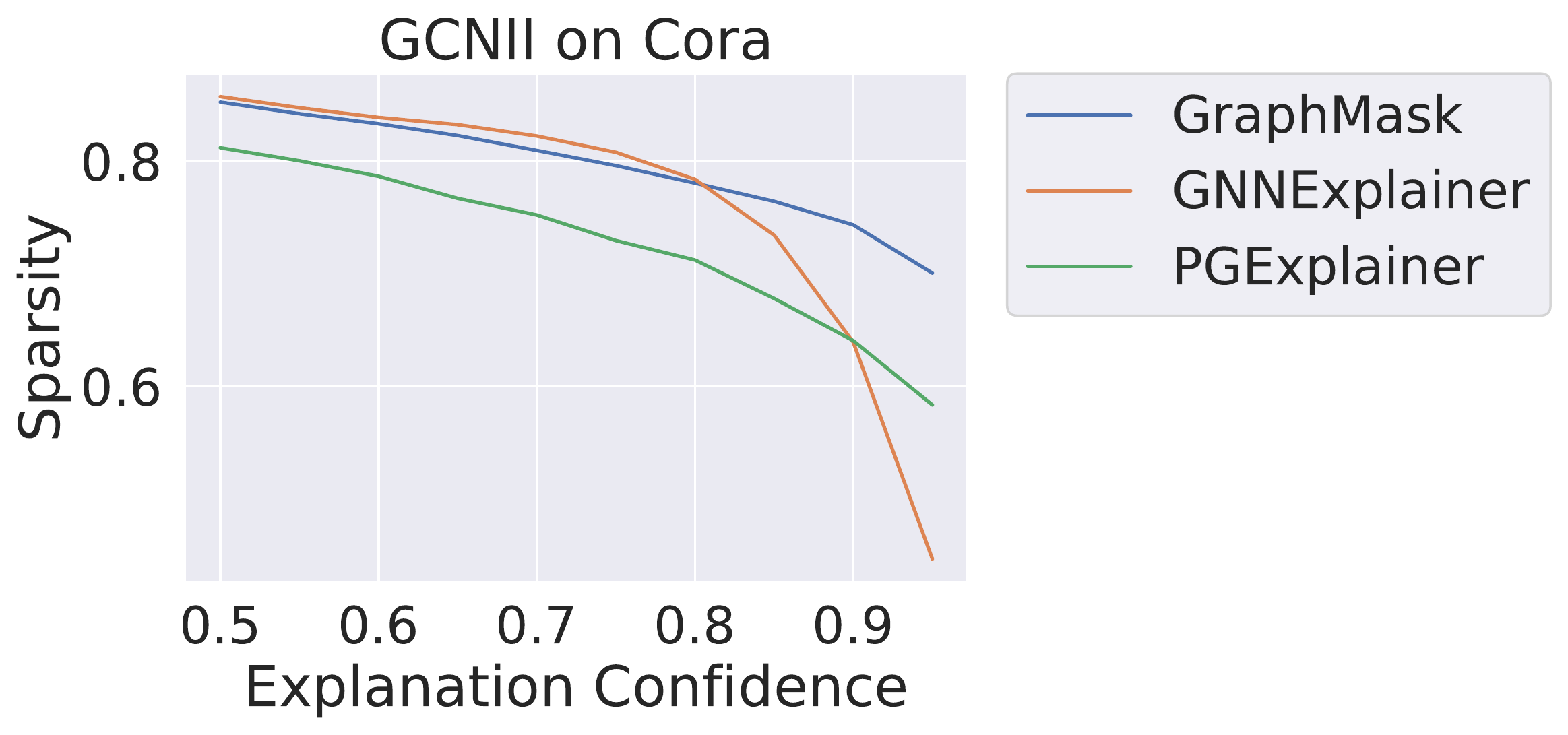}
        \end{minipage}
    \end{minipage}
    \\
    \begin{minipage}{0.97\linewidth}
        \begin{minipage}{0.32\linewidth}
            \centering
            \includegraphics[width=0.99\linewidth]{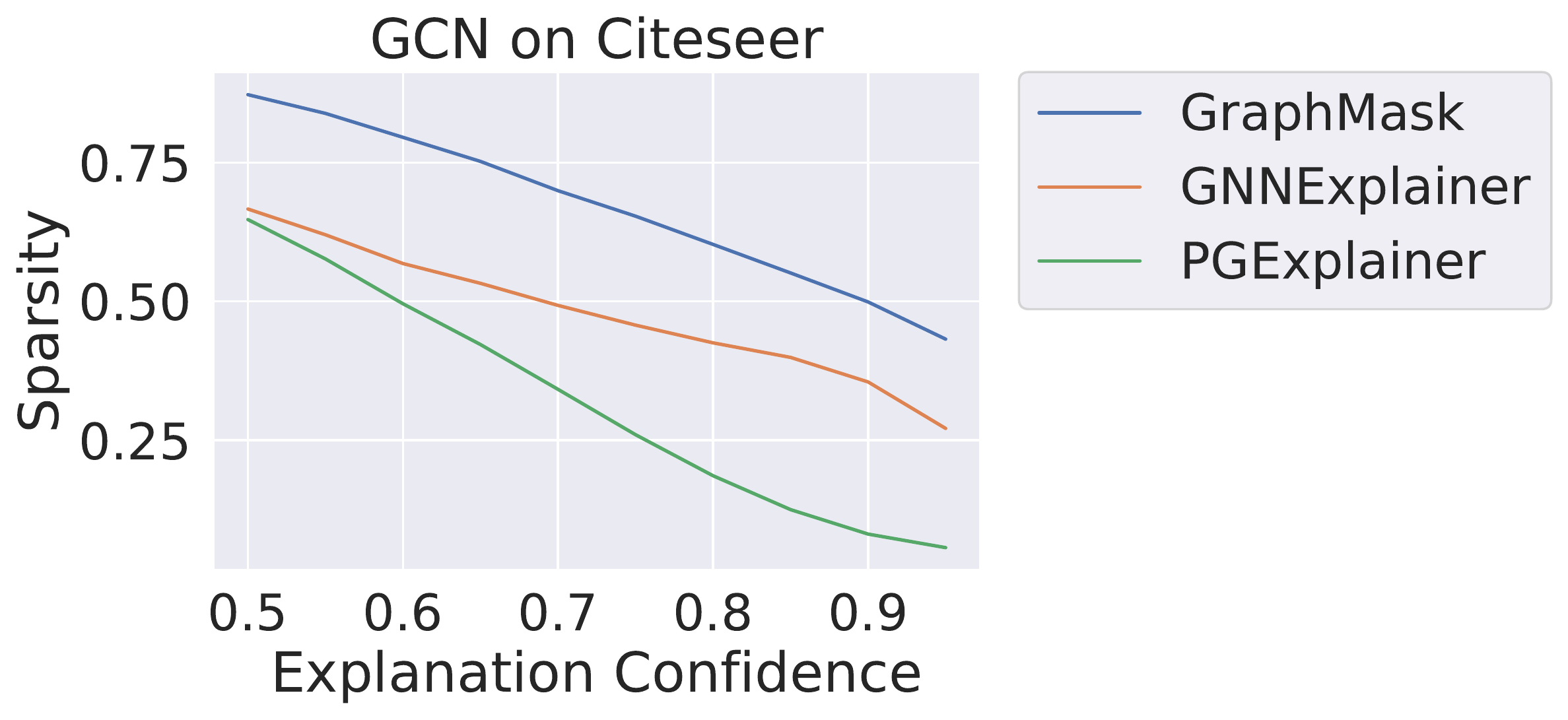}
        \end{minipage}
        \begin{minipage}{0.32\linewidth}
            \centering
            \includegraphics[width=0.99\linewidth]{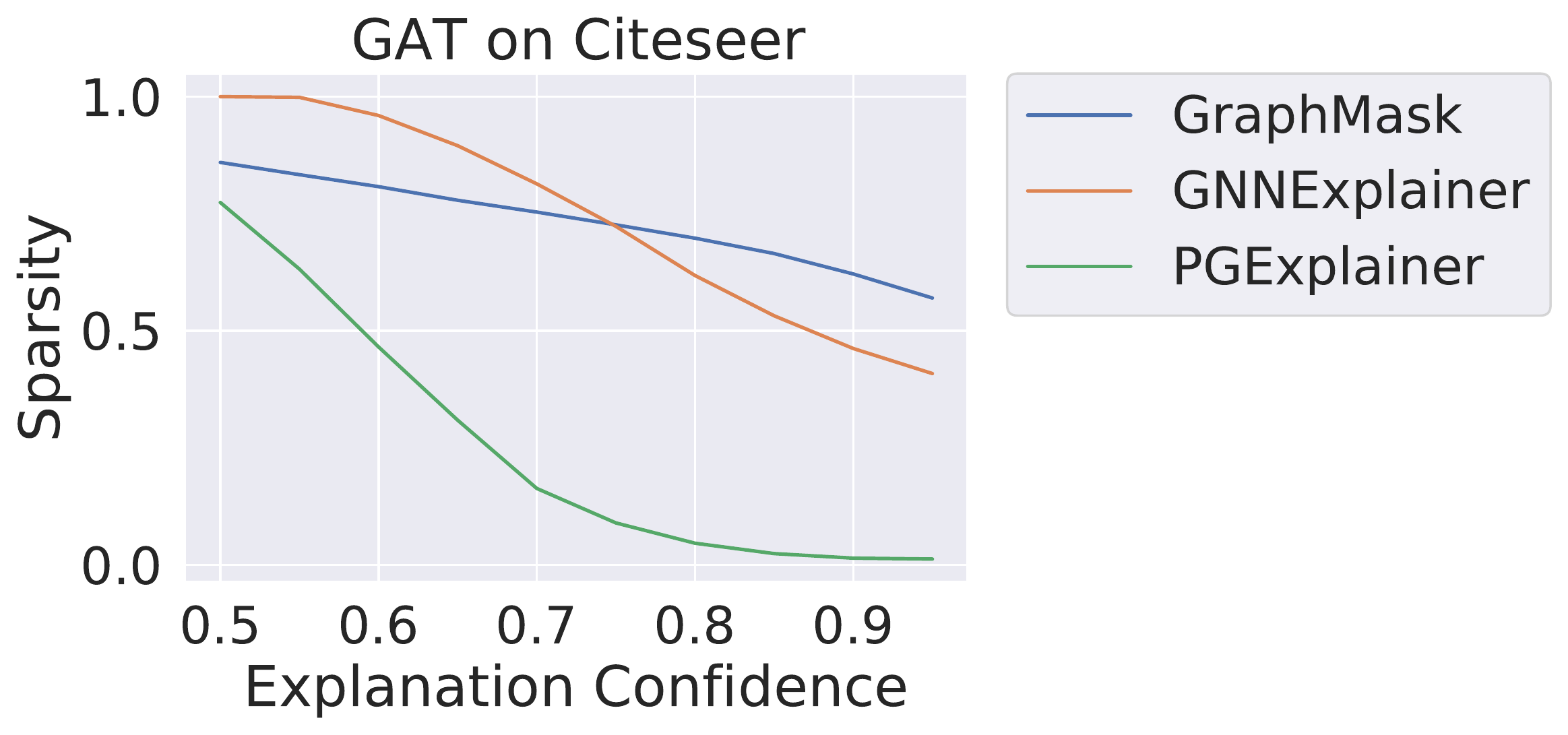}
        \end{minipage}
        \begin{minipage}{0.32\linewidth}
            \centering
            \includegraphics[width=0.99\linewidth]{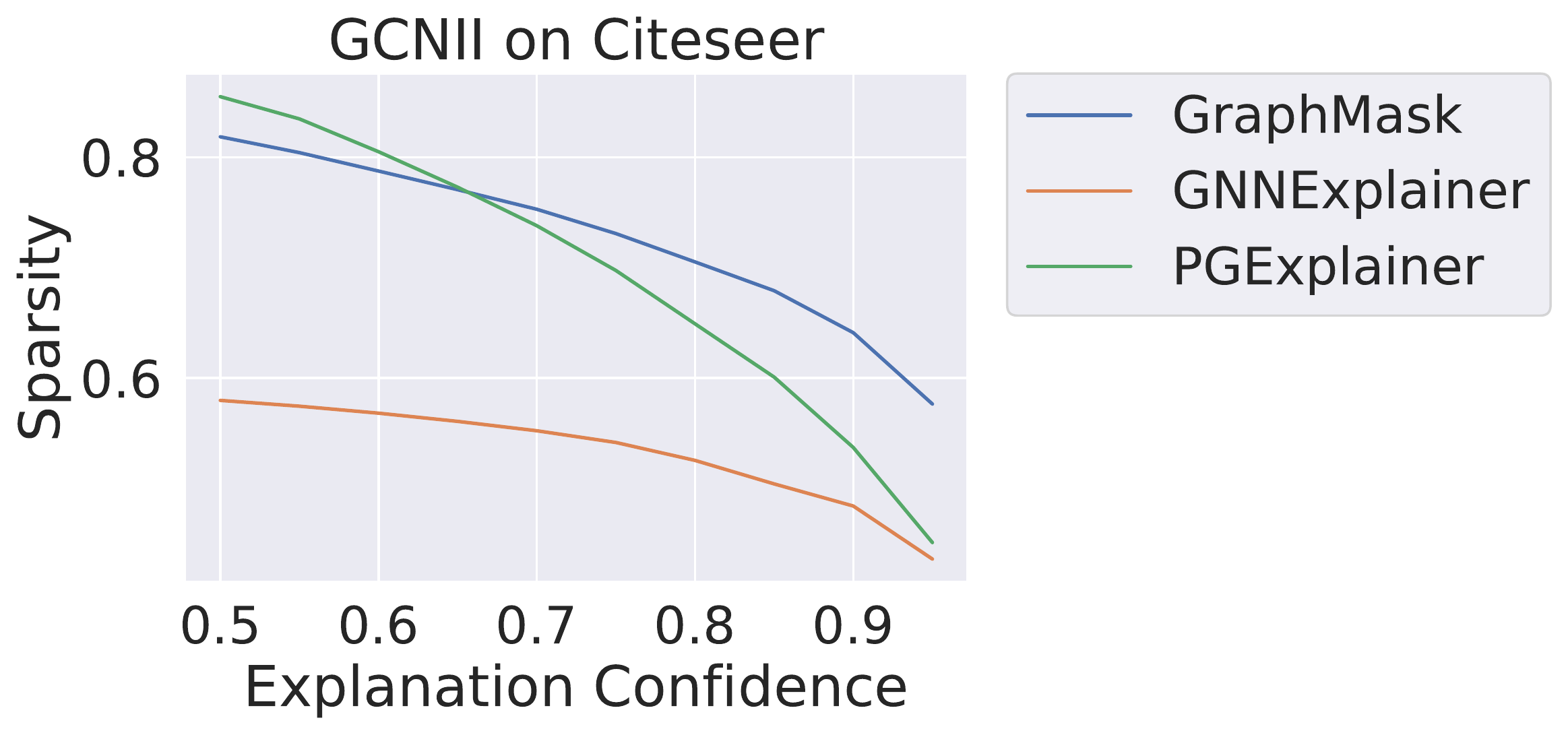}
        \end{minipage}
    \end{minipage}
    \\
    \begin{minipage}{0.97\linewidth}
        \begin{minipage}{0.32\linewidth}
            \centering
            \includegraphics[width=0.99\linewidth]{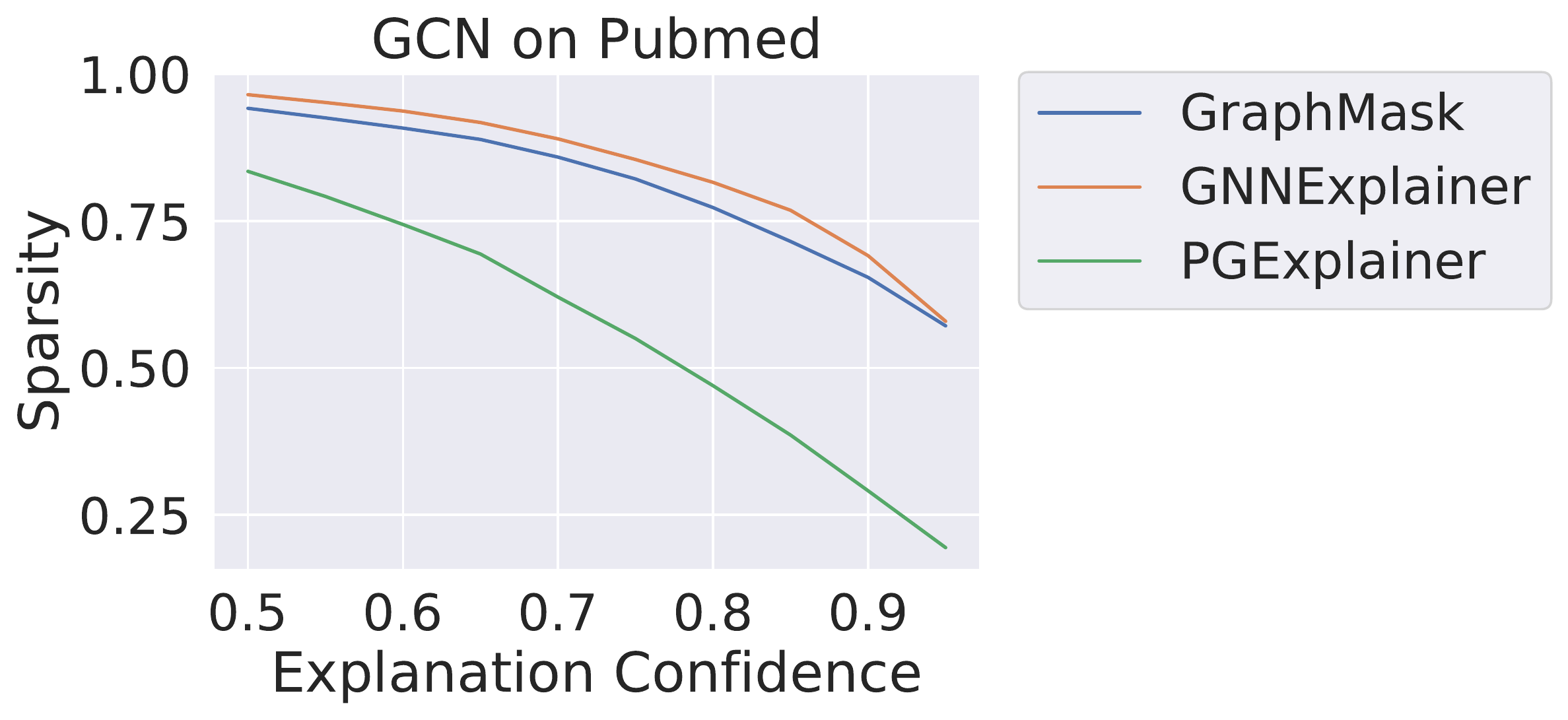}
        \end{minipage}
        \begin{minipage}{0.32\linewidth}
            \centering
            \includegraphics[width=0.99\linewidth]{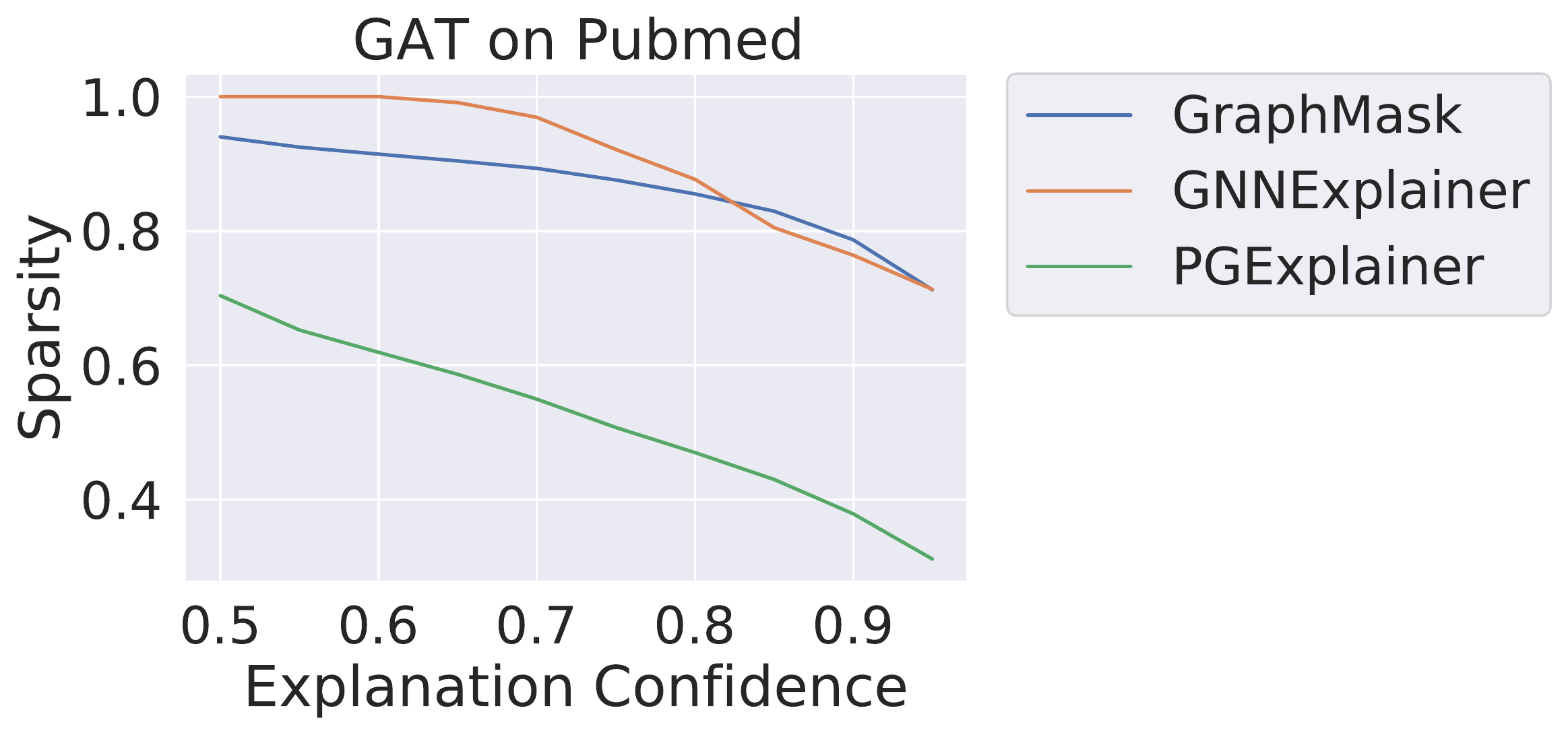}
        \end{minipage}
        \begin{minipage}{0.32\linewidth}
            \centering
            \includegraphics[width=0.99\linewidth]{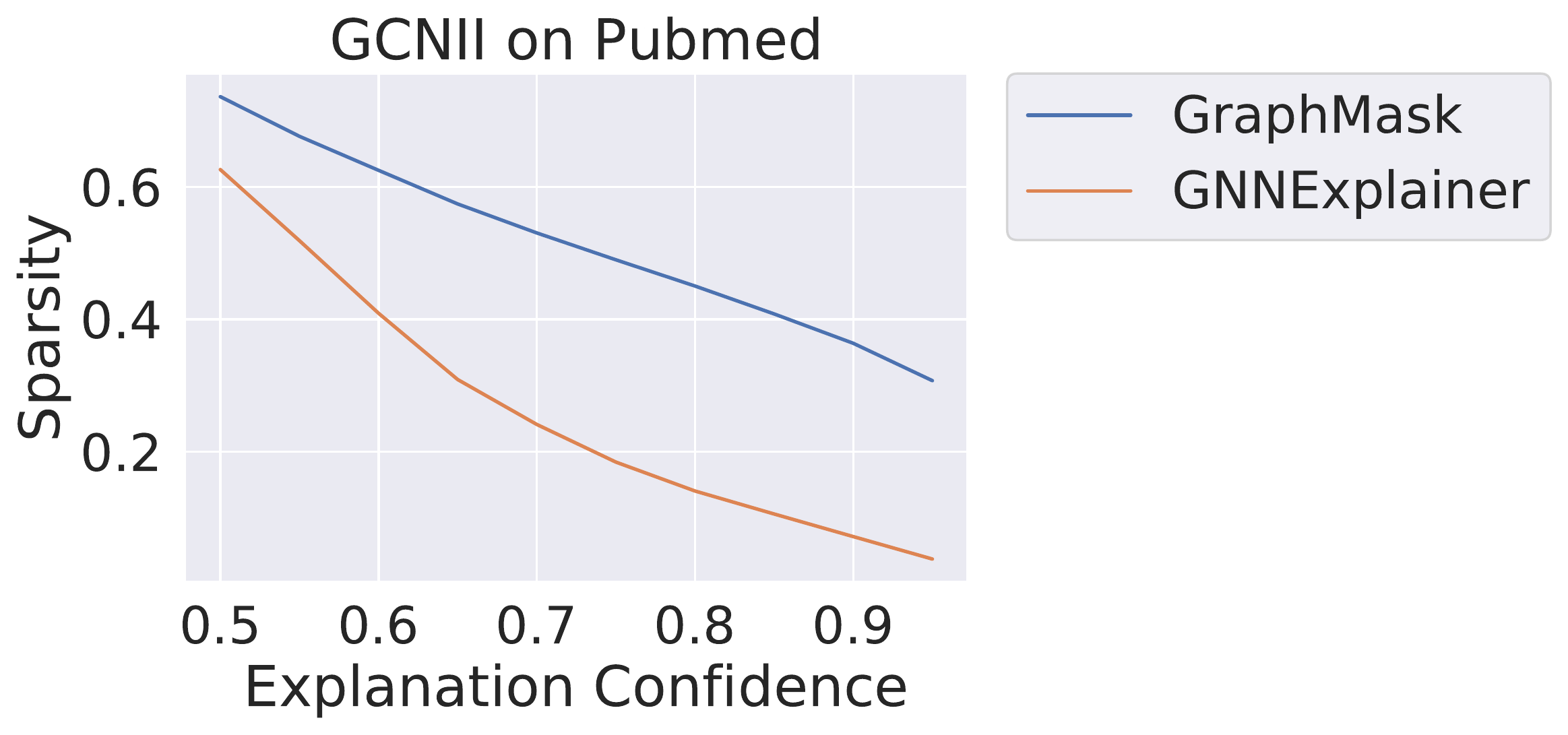}
        \end{minipage}
    \end{minipage}
    \caption{Performance of GraphMASK, GNNExplainer and PGExplainer on GCN, GAT, GCNII. No result for PGExplainer on GCNII for Pubmed because running out of memory. At the same evaluation confidence (EC), higher sparsity means a smaller number of edges are retained. A good explanation should has as high sparsity as possible with high evaluation confidence.}\label{fig:foobar}
\end{figure*}

\subsection{Datasets}
GNN explainability methods are commonly evaluated on \textbf{synthetic datasets} , where the most popular method to generate synthetic datasets is introduced in \cite{ying2019gnnexplainer}, which starts with a base graph and then attaches motifs to randomly selected nodes in the base graph. 
Specifically, they constructed the following five datasets: 
(1) \textit{BA-shapes}: It is a node classification dataset, where the base graph is a Barabási-Albert graph and the motif is a house structure. 
There are four classes in this dataset; 
one of the classes indicates the nodes in the base graph while the rest of the classes indicate the relative location of nodes in the motif. 
There is no node feature in this dataset.
(2) \textit{BA-community}:  It extends BA-shapes to more complicated scenarios with 8 classes. Basically, there are two sets of BA-shape structures, and the nodes belonging to different sets have different labels even when their relative locations within the set are the same. To distinguish the two sets, node features are introduced.
(3) \textit{Tree-Cycle}: It is a node classification dataset with two classes, where the base structure is a binary tree, and the motif is a 6-node cycle structure. The class only indicates whether the nodes are in motifs.
(4) \textit{Tree-Grids}: It is similar to Tree-Cycle, with the only difference being that the motif is a 9-node grid.
(5) \textit{BA-2Motifs}: It is a graph classification dataset with 2 classes. The graphs are obtained by attaching different motifs to Barabási-Albert graphs, and classes indicate the type of motifs in the graphs. 

In addition to synthetic datasets, \textbf{molecular datasets} MUTAG\cite{debnath1991structure} and BBBP \cite{wu2018moleculenet} are commonly used for explanations over graph classification tasks. 
In these datasets, each graph represents a molecule, where the nodes represent atoms and edges represent bonds between atoms. 
The labels for these graphs are determined by the chemical functionalities of the corresponding molecules.

%% file: experiments.tex
\section{Experiments}
\label{sec:exp}

\begin{table*}
    \small
    \begin{minipage}{\linewidth}
        \centering
        \begin{subtable}{0.5\linewidth}
          \centering
            \begin{tabular}{c|r|r|r}
                \toprule
                Method&GCN&GAT&GCNII\\
                \midrule

                \multirow{2}{*} {} GNNEx-&$\lambda=0.3912$&$\lambda=\mathbf{0.4496}$&$\lambda=0.2827$\\
                                -plainer&$\mu=-0.2726$    &$\mu=-0.2177$    &$\mu=\mathbf{0.0193}$\\\cline{1-4}
                \multirow{2}{*}{}Graph-&$\lambda=\mathbf{0.5777}$&$\lambda=0.2280$&$\lambda=\mathbf{0.6015}$\\
                                    -MASK&$\mu=\mathbf{-0.0244}$    &$\mu=\mathbf{-0.0115}$    &$\mu=-0.1605$\\\cline{1-4}
                \multirow{2}{*}{}PGEx-&$\lambda=0.2164$&$\lambda=0.1998$&$\lambda=0.1073$\\
                                 -plainer&$\mu=-0.4794$    &$\mu=-0.3494$    &$\mu=0.0027$\\
                \bottomrule
            \end{tabular}
            \caption{Results for Cora}
        \end{subtable}%
        \begin{subtable}{0.5\linewidth}
          \centering
            \begin{tabular}{c|r|r|r}
                \toprule
                Method&GCN&GAT&GCNII\\
                \midrule

                \multirow{2}{*} {} GNNEx-&$\lambda=0.1475$&$\lambda=0.3825$&$\lambda=0.1778$\\
                                -plainer&$\mu=-0.0887$    &$\mu=-0.0904$    &$\mu=\mathbf{-0.0314}$\\\cline{1-4}
                \multirow{2}{*}{}Graph-&$\lambda=\mathbf{0.3080}$&$\lambda=\mathbf{0.4556}$&$\lambda=\mathbf{0.5777}$\\
                                    -MASK&$\mu=\mathbf{-0.0341}$    &$\mu=\mathbf{-0.0121}$    &$\mu=-0.0930$\\\cline{1-4}
                \multirow{2}{*}{}PGEx-&$\lambda=0.0960$&$\lambda=0.1444$&\multirow{2}{*}{OOM}\\
                                 -plainer&$\mu=-0.1790$    &$\mu=-0.2216$    &\\
                \bottomrule
            \end{tabular}
            \caption{Results for Pubmed. OOM denotes out of memoery.}
        \end{subtable}%

    \end{minipage}
    \caption{The average fidelity and inverse fidelity with sparsities from 0.5 to 0.95. $\lambda$ and $\mu$ represent fidelity and inverse fidelity respectively.}
    \label{fidelity}
\end{table*}

In this section, we compare existing GNN explainability methods over different GNN models on real-world citation networks datasets.

\textbf{Evaluation Metric} Since we cannot get ground truth explanations for such tasks, we can only evaluate \textit{faithfulness}. To do this, for all explainability methods that only output importance scores other than the sparse subgraphs, first, we need to sparsify the continuous edge importance scores to binary explanations. A trivial way is to manually set a threshold and select edges with importance scores higher than the threshold as explanations. This however does not guarantee that the generated explanations have good quality. Then with sparsity, we can generate explanations with different sparsities and compute their fidelity values, obtaining the average fidelity value over different sparsities. However, the range of fidelity values can vary significantly for different samples and models, and hence it is not possible to directly compare the performance of an explainability method on different models based on fidelity.
What's more, what we want to evaluate is the methods' ability to find an explanation model that behaves similarly to the original model while removing unimportant edges as much as possible. Sparsity, however, focuses only on removing edges rather than evaluating the quality of explanations, and hence cannot be used as an independent variable. 

To this end,  we propose an evaluation metric \textit{explanation confidence} (EC), which is formally defined as: 
\begin{align}
    EC =& 1 - \frac{|P(Y = c|G) - P(Y = c|G_S)|}{P(Y = c|G)},
   \nonumber \\& c = \mathop{\arg \max}_{c \in C} P(Y = c|G).
\end{align}
    
\noindent This equation measures the difference between the probabilities of the predicted class using the explanation subgraph and original graph, normalized by the prediction probability using the original graph.
A higher value indicates a higher confidence that the explanation subgraph reflects the actual reasoning process of the GNN model. Compared to fidelity, EC can be used as an independent variable because its value is always $[0,1]$ for any samples, which is thus normalized across all samples and models, making it possible to directly compare the performance of different explainability methods. 
In addition, by combining EC with sparsity, we can compute the average sparsity over different qualities of explanations for all samples, which has a clearer semantic compared to existing faithfulness evaluations.

\textbf{Datasets} Current experimental studies have been limited to synthetic datasets and molecular datasets, which are not the most common applications of GNNs. To evaluate the explainability methods in more realistic scenarios, we select three of the most popular standard citation network benchmarks for GNNs: Cora \cite{mccallum2000automating}, Citeseer \cite{giles1998citeseer}and Pubmed \cite{sen2008collective}. In these datasets, nodes represent documents and edges represent citations. Node features are elements of a bag-of-word representation of a document and each node belongs to a certain class.   

 \textbf{Setup} Besides GCN \cite{kipf2016semi}, we also evaluate explainability methods on GAT \cite{DBLP:conf/iclr/VelickovicCCRLB18} and GCNII \cite{DBLP:conf/icml/ChenWHDL20}. GAT uses a different aggregation function to GCN, and GCNII is GCN with extra features which allow it to have a deeper structure. As for explainability methods, we select GNNExplainer, PGExplainer and GraphMASK because they are fundamentally similar and easier to train. PGM-Explainer and SubgraphX do not scale to big graphs and XGNN can not be applied to the node classification task. 
 For each dataset, we first train the models on the training set, and then use GNNExplainer, PGExplainer and GraphMASK to explain them on the testing set. Specifically, the GCNs and GATs all have two layers, the GCNIIs on Cora, Citeseer, Pubmed have 64, 32 and 16 layers, respectively. For each sample, we compute the highest sparsities that produce the explanations with explanation confidence at 0.50, 0.55, 0.60, 0.65, 0.70,0.75, 0.80, 0.85, 0.90, 0.95. Then we average the sparsities corresponding to the different explanation confidence levels.
 Note we do not set explanation confidence at 1, which is not practical. The experiments are conducted on an Nvidia RTX3090 GPU.
 %Note we do not set explanation confidence at 1 because at this stage, fully faithfulness is not practical. The experiments are run with an Nvidia RTX3090 GPU.

\textbf{Results} Figure 1 shows the performance of GraphMASK, GNNExplainer and PGExplainer on GCN, GAT and GCNII for the three datasets. GraphMASK consistently outperforms the other two methods, especially when the explanation confidence is high. This indicates that the explanations generated by GraphMASK are closer to the actual reasoning process of the models. This can be explained by its architecture. For example, for a two-layer GCN, edges outside of a node's 1-hop neighborhood have no impact on the model's behavior at the second layer but affect the model's behavior at the first layer. GraphMASK is capable of capturing patterns like this because it treats different layers in a graph differently while the others do not.
While PGExplainer performs the worst on GCN and GAT, it outperforms GNNExplainer on GCNII. Furthermore, PGExplainer fails to distinguish important edges at a high explanation confidence level on GAT and GCN. 
The result shows a big performance drop from GNNExplainer when the models switch from shallow models (GCNs, GATs) to deep models (GCNIIs), 
which shows that GNNExplainer is less capable of explaining models with more parameters and deeper structures. 
In addition, We report the evaluation results using existing fidelity and inverse fidelity metrics for Cora and Pubmed in Table \ref{fidelity}. Overall, GraphMASK still gives the best performance while GNNExplainer has a better fidelity for GAT in Cora and a better inverse fidelity for GCNII in Pubmed. 

For both GraphMASK and PGExplainer, we benchmark them using both single-instance level and model-level learning. The results we report for GraphMASK are using single-instance level learning, which are better than using model-level learning. The results we report for PGExplainer are using model-level learning, which are similar to using single-instance level learning. We observe that model-level learning does not give better performance in theses dataset, and the reason behind this, which is beyond the scope of this paper, can be investigated in future work.
% Investigating why model-level learning does not give better performance in these datasets can be a future 
%we provide the result using fidelity and inverse fidelity in Table \ref{fidelity}.

%% file: future.tex
\section{Future Direction}
As explainability in GNNs is still a relatively new area, there are still a lot of challenges. We suggest three future research directions as follows.

\noindent \textbf{Problem Definition} The definition of explainability is still an open question not only for explainability in GNNs but for the entire XAI community. We still need a proper and well-accepted definition of the problem that we are trying to solve: What is explainability? What do we want from explaining black-box models? What makes a good explanation? We need clearer definitions for these fundamental questions.

\noindent \textbf{Evaluation Metric} Unlike image and text domains where the evaluation can rely on human experts, graph data is difficult to visualize and difficult to understand even by human experts. Therefore, we need better quantitative evaluation metrics that are both task-agnostic and task-specific and easy to understand by general users.

\noindent \textbf{Explainable GNN for Traditional Graph Algorithms} GNNs have been used for many traditional graph mining problems, such as subgraph enumeration and counting \cite{duong2021efficient,yang2021efficient} and community computation \cite{fang2020effective,liu2019community,yang2020effective}. 
As traditional methods are based on logic and mathematics, it will be interesting to study the relationships between the explanations of GNN-based methods and the logic behind the algorithm-based methods. 

% \noindent \textbf{Interpretable Model} Although it is important to understand black-box models, it is always desirable if we have inherently interpretable models. However, interpretable models are usually less competitive compared to black-box models. Making inherently interpretable models with better performance could be a future research direction.

%% file: conclusion.tex
\section{Conclusion}
In this survey, we provide a critical review of state-of-the-art GNN explainability methods, including quantitative metrics and datasets used in the evaluation. We also propose a new evaluation metric, explanation confidence, and present our experimental results comparing various GNN explainability methods on different real-world datasets and GNN architectures. Finally, we suggest our views for future directions of GNN explainability.